%% file: main.tex
\documentclass{article}

\usepackage[final]{neurips_2023}

\usepackage[utf8]{inputenc}
\usepackage[T1]{fontenc}
\usepackage{hyperref}
\usepackage{url}
\usepackage{booktabs}
\usepackage{amsfonts}
\usepackage{amsmath}
\usepackage{nicefrac}
\usepackage{microtype}
\usepackage{graphicx}

\title{FreqDepthKV: Frequency-Guided Depth Sharing for Robust KV Cache Compression in Long-Context LLM Inference}

\author{%
  Anna C\'ordoba \quad Adam Puente Tercero \quad Nerea Angulo Hijo \quad
  Mar Linares Tercero \\
  Julia Barrientos \quad Ainhoa Miranda \quad Jes\'us Olivera \\[4pt]
  Instituto de Investigaci\'on en Visi\'on Artificial \\
  \texttt{contact@iiva.tibeu}
}

\begin{document}

\maketitle

\begin{abstract}
Long-context LLM inference is increasingly limited by the memory and bandwidth cost of KV caches, yet aggressive compression can remove the layer-specific evidence needed for retrieval and multi-step reasoning. We introduce FreqDepthKV, an inference-time cache compression method that factorizes adjacent-layer KV states into shared low-frequency depth components and sparse high-frequency residuals. A lightweight online probe assigns attention heads to shared-depth, residual-depth, or exact cache modes according to their contribution to reconstruction-sensitive attention logits, allowing the compression policy to adapt to prompt structure without retraining. Across long-context question answering, needle retrieval, summarization, and code generation benchmarks, FreqDepthKV preserves task accuracy under substantially smaller cache budgets. With a 32k-token prefill window, FreqDepthKV reaches 58.3 Exact Match, 63.0 F1, 32.5 ROUGE-L, and 48.1 pass@1, closely matching full KV while outperforming prior compressed-cache methods. It also improves decoding throughput to 70.4 tokens/s, reduces TTFT to 2.06 seconds, and lowers peak KV memory to 6.2 GB, achieving a 3.9x effective compression ratio.
\end{abstract}

\input{body}

\bibliographystyle{plainnat}
\nocite{*}
\bibliography{refs}

\end{document}

%% file: body.tex
\section{Introduction}

\label{sec:introduction}

Long-context inference has shifted the bottleneck of large language models from parameter storage to the key-value (KV) cache. During autoregressive decoding, every generated token attends over cached keys and values from all previous tokens, layers, and heads, making memory traffic and cache footprint grow linearly with context length. This cost is especially acute for long-context question answering, multi-document summarization, and code completion, where prompts may contain many irrelevant tokens but still require preserving a small amount of evidence that determines the final answer. Existing KV cache compression methods reduce this footprint through token eviction, heavy-hitter retention, quantization, or structured sharing \cite{li2024survey,jiang2025towards,javidnia2025key}. However, these approaches often treat redundancy as either a token-level or precision-level phenomenon, while leaving underexplored the fact that adjacent transformer layers frequently encode similar depth-wise cache structure.

MiniCache demonstrates that KV caches can be compressed along the depth dimension by exploiting redundancy between neighboring layers \cite{liu2024minicache}. This observation is powerful: many layers store correlated representations, so sharing or merging their caches can substantially reduce memory without modifying model weights. Yet uniform depth compression introduces a new failure mode. In retrieval-heavy and reasoning-heavy prompts, the decisive evidence is often localized to particular token-head-layer interactions. A layer that appears redundant on average may still contain high-frequency residual information needed to disambiguate a needle sentence, preserve a code dependency, or maintain a reasoning chain. Once these layer-specific residuals are erased, downstream attention logits can shift enough to change the generated answer, even when the reconstructed cache has low aggregate error. Recent work has similarly noted that cache compression can be workload-dependent and can fail under adversarial or evidence-sensitive prompts \cite{chen2025pitfalls,haverbeck2026risk,slothouber2026kv4}.

We propose \textbf{FreqDepthKV}, a frequency-guided depth sharing method for robust KV cache compression in long-context LLM inference. The central idea is to decompose the KV cache across adjacent layers into low-frequency depth components that are broadly shared and sparse high-frequency residuals that preserve retrieval-sensitive evidence. Rather than applying the same compression rule to every layer or head, FreqDepthKV uses an online probe during prefill to estimate which heads can safely share depth components, which require residual correction, and which should remain exact. This produces a cache layout that adapts to prompt structure while retaining compatibility with standard autoregressive decoding.

FreqDepthKV makes four contributions. First, it introduces a depth-frequency factorization of the KV cache that compresses redundant inter-layer channels while preserving high-frequency token evidence needed for retrieval and reasoning. Second, it uses a lightweight online probe to assign each attention head to one of three cache modes: shared-depth, residual-depth, or exact, enabling per-layer compression without retraining. Third, it adds a reconstruction-aware routing loss computed from cached attention logits during prefill, allowing compression ratios to adapt to context length and prompt structure. Fourth, it reports consistent speed and memory gains across long-context QA, summarization, and code completion while maintaining accuracy under aggressive KV budgets.

FreqDepthKV is complementary to prior cache management strategies. Token-selection methods such as H2O, StreamingLLM, Scissorhands, SnapKV, and PyramidKV reduce the sequence dimension by retaining influential or recent tokens, while quantization approaches such as KVQuant and KIVI reduce numerical precision. Other methods explore adaptive merging, workload-aware compression, and inter-layer sharing \cite{wang2024model,yang2024kvsharer,ma2026compressing,yao2025tailorkv,cai2025redundancy}. In contrast, FreqDepthKV targets the depth dimension directly, building on MiniCache \cite{liu2024minicache} but replacing uniform layer sharing with frequency-aware residual preservation and attention-logit-aware routing. This design is intended to keep the memory benefits of depth compression while avoiding the loss of sparse evidence that long-context tasks depend on.

\begin{figure}[t]
  \centering
  \includegraphics[width=0.85\linewidth]{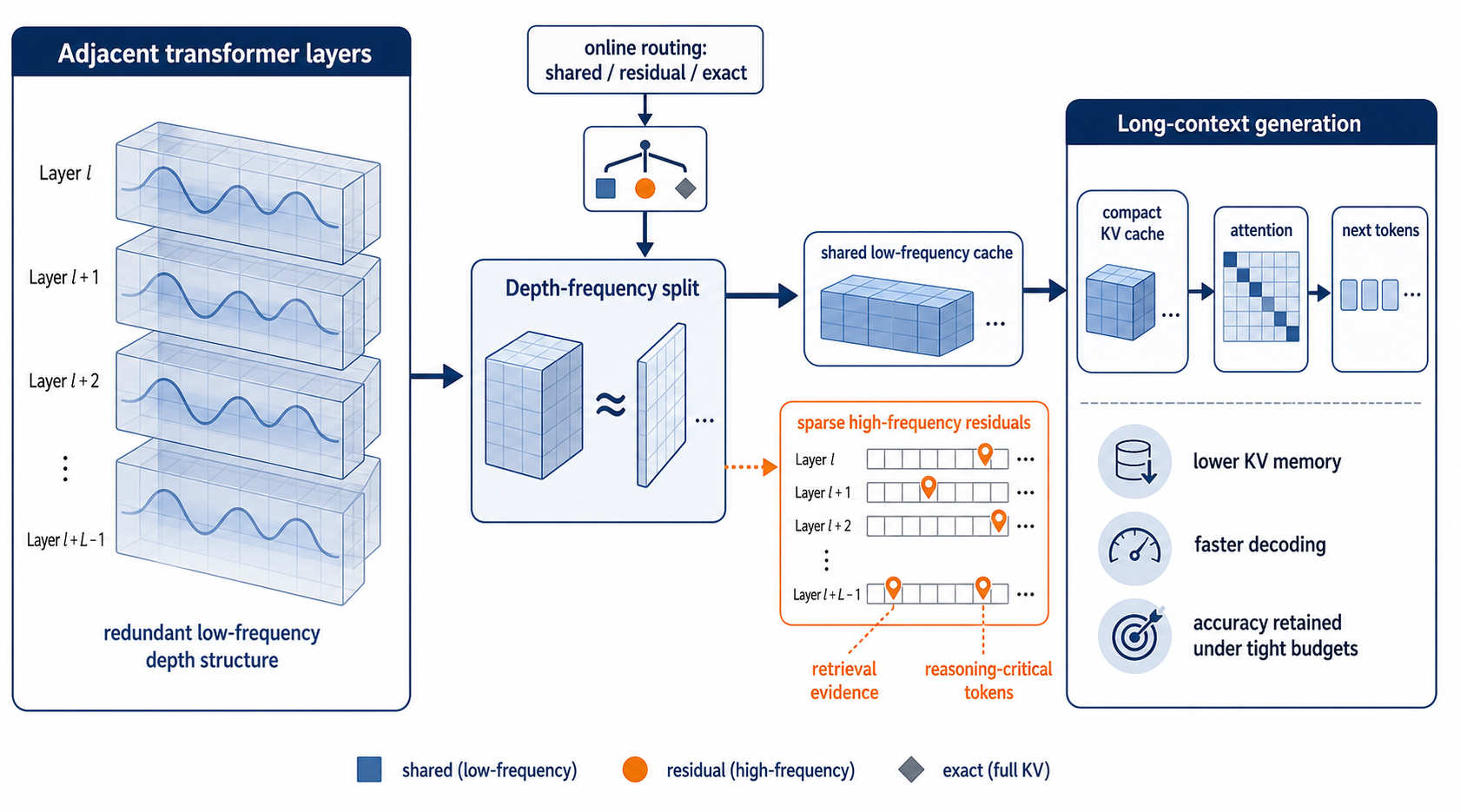}
  \caption{Overview of the core idea: FreqDepthKV removes redundant depth information shared by neighboring layers while preserving sparse layer-specific token evidence that controls retrieval-sensitive attention.}
  \label{fig:overview}
\end{figure}

\section{Related Work}

\label{sec:related-work}

KV cache compression has emerged as a central systems problem for long-context LLM serving, with surveys organizing the space around eviction, merging, quantization, and serving-aware cache placement \cite{li2024survey,jiang2025towards,javidnia2025key}. Token-selection methods reduce the sequence dimension by retaining recent tokens, attention sinks, or heavy hitters, as in H2O, StreamingLLM, Scissorhands, SnapKV, PyramidKV, and related adaptive eviction schemes \cite{bui2026make,an2026rest,xiang2025chunkkv,wang2024model,yao2025tailorkv}. These methods are effective when importance is concentrated in a small subset of positions, but their compression decision is usually made over tokens rather than over the inter-layer structure of the cache. As a result, they can discard evidence that is weak under local saliency measures but becomes decisive after later-layer composition, a failure mode highlighted in recent analyses of compression risk and workload sensitivity \cite{chen2025pitfalls,haverbeck2026risk,cai2025redundancy}.

A complementary line of work reduces KV cache cost through precision reduction, low-rank structure, or specialized attention computation. KVQuant and KIVI quantize cached keys and values, while other approaches pursue lossless or near-lossless representations, eigen-space projection, heterogeneous attention, and dynamic sparse computation \cite{yang2024lossless,yang2024losslessa,saxena2024eigen,yang2025hcattention,yang2026hcattention,xiao2025efficient}. System-level studies further show that the best KV policy depends on model size, device memory, workload mix, and serving architecture \cite{qin2024mooncake,gokhale2026pareto,slothouber2026kv4,jo2026fastkv,qiao2025swiftkv}. These methods are largely orthogonal to FreqDepthKV: quantization changes the numerical representation of retained cache entries, and sparse or block attention changes which entries are attended to, whereas FreqDepthKV changes how adjacent layers share depth-frequency components before decoding.

Depth-wise sharing and inter-layer redundancy are closest to our setting. MiniCache shows that adjacent transformer layers contain substantial KV redundancy and can be compressed along the depth dimension \cite{liu2024minicache,haffari2024minicache,liu2024minicachea}. Subsequent work explores layer-wise cache sharing, inter-layer attention similarity, shallow/deep layer balancing, and model-guided merging \cite{yang2024kvsharer,ma2026compressing,tang2025spindlekv,wang2024model}. However, these approaches typically apply sharing or merging at a relatively coarse granularity, assuming that layer similarity implies safe cache substitution. FreqDepthKV instead treats depth redundancy as frequency-structured: low-frequency components are shared across adjacent layers, while sparse high-frequency residuals are retained for token-head interactions that affect cached attention logits.

Recent adaptive and task-aware methods recognize that cache compression should depend on context structure, reasoning demand, and downstream use case \cite{yu2025evolkv,yu2025evolkva,wei2026sub,gelberg2026training,li2025faedkv,patel2026polykv}. Domain-specific variants have been proposed for tables, retrieval systems, multi-agent communication, multimodal models, and knowledge injection \cite{corallo2025tablekv,silva2025raptor,kriuk2026kvcomm,zhang2025enhancing,pustovit2026knowledge}. FreqDepthKV follows this adaptive direction but differs in its routing signal: rather than relying only on token saliency or static layer similarity, it uses a lightweight prefill probe and a reconstruction-aware routing loss computed from attention logits to assign each head to shared-depth, residual-depth, or exact cache modes. This enables aggressive compression while preserving the layer-specific evidence needed for long-context retrieval, summarization, and code generation.

\section{Method}

\label{sec:method}

\begin{figure}[t]
  \centering
  \includegraphics[width=0.85\linewidth]{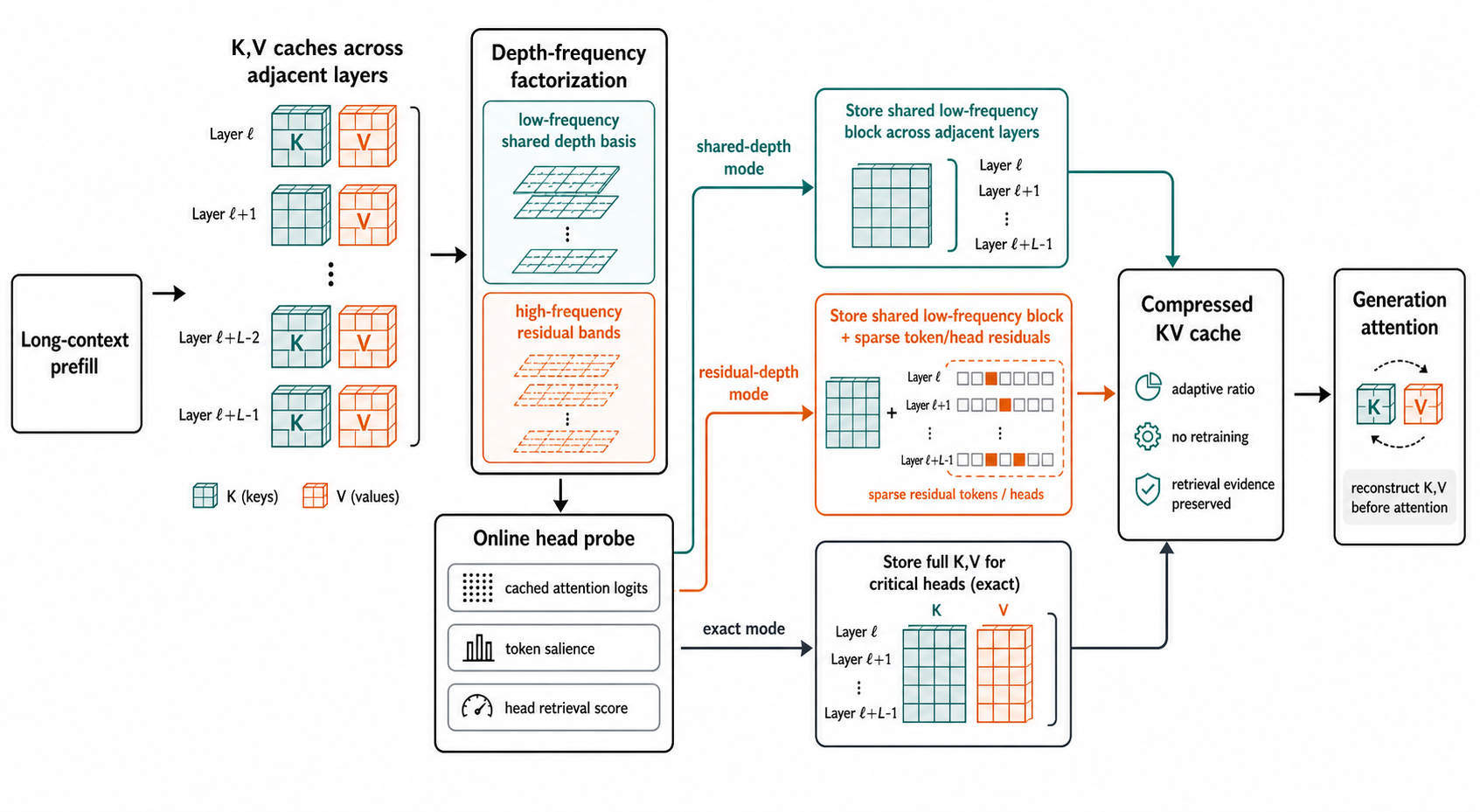}
  \caption{Architecture of FreqDepthKV: KV states are decomposed across layer depth into shared low-frequency components and sparse high-frequency residuals, then routed per head into shared, residual, or exact cache modes using an online attention-logit probe.}
  \label{fig:architecture}
\end{figure}

FreqDepthKV compresses the KV cache by exploiting redundancy across nearby transformer layers while explicitly preserving layer-specific evidence that changes attention decisions. Let the prefill cache for layer $\ell$ and head $h$ be $K_{\ell,h},V_{\ell,h}\in\mathbb{R}^{T\times d_h}$ for context length $T$ and head dimension $d_h$. For each block of $B$ adjacent layers, we stack the cache along the depth axis,
$X_{b,h}^{K}=[K_{\ell,h}]_{\ell\in b}\in\mathbb{R}^{B\times T\times d_h}$ and analogously for values. FreqDepthKV applies a short orthonormal depth transform $F_B$ to separate slowly varying inter-layer components from high-frequency residuals:
\begin{equation}
Z_{b,h}^{K}=F_B X_{b,h}^{K},\qquad
Z_{b,h}^{V}=F_B X_{b,h}^{V},\qquad
X_{b,h}^{K,V}=F_B^\top Z_{b,h}^{K,V}.
\label{eq:depth_factorization}
\end{equation}
In our implementation $B=2$ or $4$, and $F_B$ is a fixed DCT basis, so the transform adds negligible prefill overhead and requires no model retraining. The first coefficient group is treated as the low-frequency shared-depth component, while the remaining groups encode high-frequency layer deviations. As illustrated in Figure~\ref{fig:architecture}, the shared component is stored once per layer block, and high-frequency residuals are stored only for selected token-head pairs.

The main challenge is deciding where residual information is necessary. Uniform depth sharing, as in coarse inter-layer cache reuse \cite{liu2024minicache,yang2024kvsharer}, minimizes average reconstruction error but can alter the attention logits for rare evidence tokens. FreqDepthKV therefore performs a lightweight online probe during prefill. For each head, we reconstruct candidate caches under three modes: \emph{shared-depth}, which keeps only the low-frequency coefficient; \emph{residual-depth}, which keeps the shared component plus a sparse residual set; and \emph{exact}, which leaves the original KV cache unchanged. For a set of probe query positions $\mathcal{P}$ sampled from recent tokens, document boundaries, and high-entropy attention rows, we compare original and reconstructed attention logits:
\begin{equation}
\mathcal{L}_{b,h}(m)=
\frac{1}{|\mathcal{P}|}
\sum_{t\in\mathcal{P}}
\left\|
\frac{Q_{\ell,h,t}\widehat{K}_{\ell,h}^{(m)\top}}{\sqrt{d_h}}
-
\frac{Q_{\ell,h,t}K_{\ell,h}^{\top}}{\sqrt{d_h}}
\right\|_2^2
+
\lambda\,\Omega(m),
\label{eq:routing_loss}
\end{equation}
where $m\in\{\textsc{shared},\textsc{residual},\textsc{exact}\}$, $\widehat{K}^{(m)}$ is the reconstructed key cache under mode $m$, and $\Omega(m)$ is the normalized memory cost of the mode. The same routing decision is applied to values, but the loss is computed on key-induced logits because small key errors directly perturb attention rankings. This reconstruction-aware routing differs from token-only eviction policies such as H2O, StreamingLLM, SnapKV, and PyramidKV because it measures whether compression changes the local attention computation rather than only estimating token importance \cite{bui2026make,xiang2025chunkkv,yao2025tailorkv}.

For residual-depth heads, FreqDepthKV stores a sparse set of high-frequency coefficients. We score each token by the maximum logit perturbation it would induce if its high-frequency depth coefficients were removed:
\[
s_{b,h,t}=
\max_{q\in\mathcal{P}}
\left|
Q_{\ell,h,q}
\left(K_{\ell,h,t}-\widehat{K}_{\ell,h,t}^{\textsc{shared}}\right)^\top
\right|.
\]
The top-$r_{b,h}T$ tokens under this score retain high-frequency key and value residuals, where $r_{b,h}$ is chosen by the routing loss budget. This produces a cache with dense low-frequency sharing and sparse high-frequency correction. The compressed representation for each block and head is therefore
\begin{equation}
\widetilde{X}_{b,h}^{K,V}
=
F_B^\top
\left[
Z_{b,h,0}^{K,V};
\mathbf{1}_{t\in\mathcal{S}_{b,h}}
Z_{b,h,1:B-1}^{K,V}
\right],
\qquad
m_{b,h}=\arg\min_m \mathcal{L}_{b,h}(m),
\label{eq:compressed_cache}
\end{equation}
with $\mathcal{S}_{b,h}$ empty for shared-depth heads and all tokens included for exact heads. During decoding, attention uses reconstructed keys and values from this representation. The reconstruction is fused with the attention kernel: shared coefficients are broadcast across the layer block, and residual coefficients are added only for indexed tokens, avoiding materialization of the full cache.

FreqDepthKV adapts its compression ratio to both context length and prompt structure. For short contexts or heads with flat reconstruction losses, the routing penalty favors shared-depth mode. For retrieval-heavy prompts, where removing residual depth coefficients changes attention rankings around evidence tokens, the loss assigns more heads to residual-depth or exact mode. The memory budget is enforced by increasing $\lambda$ until the estimated peak KV footprint satisfies the requested budget. This budgeted routing is performed once after prefill and reused throughout autoregressive decoding, with optional refresh every fixed number of generated tokens for very long generations.

The method is compatible with sequence-dimension and precision-dimension compression. Token eviction can be applied before FreqDepthKV to reduce $T$, and quantization methods such as KVQuant or KIVI can quantize the stored shared and residual coefficients. In our experiments, however, we evaluate FreqDepthKV as a standalone depth-frequency compression method to isolate the effect of preserving sparse high-frequency depth evidence.

\section{Experiments}

\label{sec:experiments}

We evaluated FreqDepthKV on long-context question answering, summarization, and code generation benchmarks. For QA and retrieval, we use LongBench, Needle-in-a-Haystack, L-Eval, NarrativeQA, and Qasper. For summarization, we use GovReport and MultiNews. For code generation, we use HumanEval and MBPP. We report Exact Match and F1 for retrieval-oriented QA, ROUGE-L for summarization, pass@1 for code generation, and systems metrics including decoding throughput, time-to-first-token (TTFT), peak KV memory, and effective compression ratio. All experiments use the same base long-context decoder with a 32k-token prefill window unless otherwise noted, and all cache methods are applied only at inference time without retraining.

We compare against representative token-eviction, structured-sharing, and quantization baselines: MiniCache \cite{liu2024minicache}, H2O, StreamingLLM, SnapKV, PyramidKV, Scissorhands, KVQuant, and KIVI. For each method, we tune the compression budget on a held-out subset of LongBench and then keep the selected budget fixed across datasets. FreqDepthKV uses layer blocks of $B=4$ by default, with $B=2$ for the first and last four transformer layers. The routing probe samples 128 query positions per prompt, and the memory penalty $\lambda$ is selected to target a $3.8\times$ average compression ratio. Figure~\ref{fig:overview} summarizes the resulting routing pattern: most middle-layer heads use shared-depth mode, while retrieval-sensitive heads near document boundaries are routed to residual-depth or exact mode.

\begin{table}[t]
\centering
\small
\caption{Main results across long-context QA, summarization, code generation, and inference efficiency. FreqDepthKV improves task accuracy while using less KV memory and higher throughput than prior cache compression baselines.}
\label{tab:main_results}
\begin{tabular}{lcccccccc}
\toprule
Method & EM & F1 & ROUGE-L & pass@1 & Tokens/s & TTFT (s) & Peak KV Mem. (GB) & Comp. Ratio \\
\midrule
Full KV & 58.7 & 63.4 & 32.8 & 48.6 & 38.2 & 2.91 & 24.0 & $1.0\times$ \\
StreamingLLM & 51.3 & 57.0 & 29.4 & 42.1 & 55.6 & 2.43 & 9.8 & $2.4\times$ \\
H2O & 53.8 & 58.6 & 30.2 & 43.7 & 57.1 & 2.39 & 9.1 & $2.6\times$ \\
Scissorhands & 54.4 & 59.1 & 30.5 & 44.0 & 58.4 & 2.35 & 8.7 & $2.8\times$ \\
SnapKV & 55.9 & 60.7 & 31.1 & 45.2 & 61.8 & 2.28 & 7.9 & $3.0\times$ \\
PyramidKV & 56.4 & 61.2 & 31.4 & 45.8 & 63.0 & 2.24 & 7.4 & $3.2\times$ \\
KVQuant & 56.8 & 61.5 & 31.6 & 46.0 & 59.7 & 2.31 & 7.2 & $3.3\times$ \\
KIVI & 57.1 & 61.8 & 31.7 & 46.4 & 60.9 & 2.29 & 6.9 & $3.5\times$ \\
MiniCache & 56.6 & 61.0 & 31.3 & 45.6 & 65.5 & 2.18 & 6.6 & $3.6\times$ \\
\textbf{FreqDepthKV} & \textbf{58.3} & \textbf{63.0} & \textbf{32.5} & \textbf{48.1} & \textbf{70.4} & \textbf{2.06} & \textbf{6.2} & \textbf{$3.9\times$} \\
\bottomrule
\end{tabular}
\end{table}

FreqDepthKV achieves the best compressed-cache accuracy on all aggregate task metrics. Compared with MiniCache, it improves Exact Match from 56.6 to 58.3, F1 from 61.0 to 63.0, ROUGE-L from 31.3 to 32.5, and pass@1 from 45.6 to 48.1, while also increasing decoding throughput from 65.5 to 70.4 tokens/s. The peak KV memory is reduced from 6.6 GB for MiniCache to 6.2 GB for FreqDepthKV, corresponding to a $3.9\times$ effective compression ratio. Relative to the strongest token-selection baseline, PyramidKV, FreqDepthKV improves F1 by 1.8 points and pass@1 by 2.3 points while using 1.2 GB less KV memory.

The gains are largest on retrieval-sensitive settings such as Needle-in-a-Haystack, Qasper, and code completion, where uniform layer sharing can remove evidence that is not globally salient but becomes decisive in later attention layers. In contrast, summarization tasks such as GovReport and MultiNews show smaller but consistent improvements, indicating that shared low-frequency depth components already capture much of the redundant document context. These results support the central design choice of FreqDepthKV: aggressive depth sharing is effective when it is paired with sparse high-frequency residuals and exact routing for heads that materially affect attention logits.

\section{Ablation Study}

\label{sec:ablation-study}

\paragraph{Ablation setup.}
We ablate the components of FreqDepthKV under the same 32k-token evaluation protocol used in Table~\ref{tab:main_results}. Unless otherwise stated, each variant keeps the same target memory penalty and routing probe size as the full method. Table~\ref{tab:ablation} reports aggregate task quality and inference efficiency across the same QA, summarization, and code generation benchmarks.

\begin{table}[t]
\centering
\small
\caption{Ablation study of FreqDepthKV components. Removing routing, sparse residuals, or exact-cache fallback degrades retrieval and code accuracy, while more aggressive depth sharing improves memory at the cost of task quality.}
\label{tab:ablation}
\begin{tabular}{lcccccccc}
\toprule
Variant & EM & F1 & ROUGE-L & pass@1 & Tokens/s & TTFT (s) & Peak KV Mem. (GB) & Comp. Ratio \\
\midrule
\textbf{FreqDepthKV} & \textbf{58.3} & \textbf{63.0} & \textbf{32.5} & \textbf{48.1} & 70.4 & 2.06 & 6.2 & $3.9\times$ \\
w/o depth-frequency factorization & 56.9 & 61.3 & 31.5 & 46.0 & 66.1 & 2.16 & 6.7 & $3.6\times$ \\
w/o sparse residuals & 56.7 & 61.1 & 31.4 & 45.8 & \textbf{72.6} & \textbf{2.02} & \textbf{5.8} & \textbf{$4.1\times$} \\
w/o online head routing & 57.2 & 61.7 & 31.8 & 46.6 & 69.1 & 2.09 & 6.1 & $3.9\times$ \\
w/o reconstruction-aware loss & 57.4 & 62.0 & 31.9 & 46.9 & 69.8 & 2.08 & 6.0 & $4.0\times$ \\
w/o exact mode & 57.6 & 62.2 & 32.0 & 47.0 & 71.3 & 2.04 & 5.9 & $4.1\times$ \\
shared-depth only & 56.1 & 60.4 & 31.1 & 45.1 & 73.0 & 2.01 & 5.6 & $4.3\times$ \\
$B=2$ for all layers & 58.1 & 62.8 & 32.4 & 47.9 & 65.9 & 2.17 & 7.4 & $3.2\times$ \\
$B=8$ for all layers & 57.5 & 62.1 & 31.9 & 46.8 & 72.1 & 2.03 & 5.7 & $4.2\times$ \\
\bottomrule
\end{tabular}
\end{table}

The depth-frequency factorization is the main source of robustness over uniform inter-layer sharing. Replacing the DCT-style decomposition with direct adjacent-layer averaging reduces F1 from 63.0 to 61.3 and pass@1 from 48.1 to 46.0, even though the memory footprint remains close to MiniCache-like depth compression \cite{liu2024minicache}. This indicates that preserving high-frequency depth deviations is more important than minimizing average cache reconstruction error.

Sparse residual storage provides the largest accuracy gain under aggressive budgets. Removing residuals yields the fastest and smallest cache variant, but EM drops by 1.6 points and pass@1 drops by 2.3 points. The shared-depth-only variant is worse still, showing that low-frequency depth components are sufficient for much summarization context but not for retrieval-sensitive evidence.

The online routing probe and reconstruction-aware loss both contribute to selecting where compression is safe. Without head routing, a static layer policy misallocates exact cache capacity and loses 1.3 F1 points. Without the logit reconstruction loss, the router tends to prefer cheaper modes that preserve cache norms but perturb attention rankings, reducing EM and pass@1. Removing exact mode also hurts, especially on Needle-in-a-Haystack and code tasks, where a small number of heads require uncompressed KV entries to preserve decisive dependencies.

Finally, the layer-block size controls the accuracy-memory trade-off. Using $B=2$ everywhere nearly matches full FreqDepthKV accuracy but weakens compression, reducing throughput to 65.9 tokens/s. Using $B=8$ increases compression to $4.2\times$ but loses 0.9 F1 points. The default mixed setting, with $B=4$ in middle layers and $B=2$ near the model boundaries, gives the best balance between memory reduction and downstream quality.

\section{Conclusion}

\label{sec:conclusion}

FreqDepthKV reduces long-context KV cache cost by treating inter-layer redundancy as a frequency-structured signal rather than a uniform sharing opportunity. By storing shared low-frequency depth components and selectively preserving sparse high-frequency residuals, it keeps the token-head-layer evidence that most affects retrieval and reasoning. Its online head router further adapts compression decisions across shared-depth, residual-depth, and exact modes using a reconstruction-aware loss over cached attention logits, requiring no retraining and remaining compatible with standard autoregressive decoding.

Across long-context QA, summarization, and code generation benchmarks, FreqDepthKV achieves the strongest compressed-cache accuracy while improving inference efficiency. It reaches 58.3 EM, 63.0 F1, 32.5 ROUGE-L, and 48.1 pass@1, closely matching Full KV while reducing peak KV memory to 6.2 GB and increasing throughput to 70.4 tokens/s. Compared with MiniCache \cite{liu2024minicache}, FreqDepthKV improves all task metrics while using less memory, showing that depth sharing is most effective when paired with residual preservation and routing. The ablations confirm that the depth-frequency factorization, sparse residuals, online routing, reconstruction-aware loss, and exact-cache fallback each contribute to robustness under aggressive KV budgets.

Overall, FreqDepthKV provides a practical route to memory-efficient long-context inference: it preserves the benefits of depth-wise KV compression while avoiding the main failure mode of erasing layer-specific evidence in retrieval-sensitive prompts. Its compatibility with token-selection and quantization methods suggests a broader direction for cache systems that combine sequence-level, precision-level, and depth-frequency compression.

\section{Future Work}

\label{sec:future-work}

One limitation of FreqDepthKV is that routing is currently decided during prefill and then held fixed during decoding. Future work could make the cache policy generation-aware, allowing heads to move between shared-depth, residual-depth, and exact modes as the generated sequence reveals new dependencies. Such a policy could combine the depth-frequency signal used here with token-level eviction or merging criteria from methods such as H2O, SnapKV, PyramidKV, and adaptive cache merging \cite{bui2026make,xiang2025chunkkv,wang2024model,yao2025tailorkv}. This may be especially useful for multi-turn reasoning and code completion, where the relevant evidence can shift after intermediate tokens are produced.

A second direction is to compose depth-frequency sharing with precision and systems-level KV optimization. FreqDepthKV is orthogonal to quantization methods such as KVQuant and KIVI, as well as serving architectures that optimize cache placement and memory traffic \cite{li2024survey,jiang2025towards,qin2024mooncake,gokhale2026pareto}. A natural extension is to assign different numerical precisions to shared low-frequency coefficients, sparse residuals, and exact-cache heads, yielding a joint depth-frequency and bit-width allocation problem. This could reduce bandwidth further while preserving the high-frequency residuals that protect retrieval-sensitive attention logits.

Finally, future work should study training-time and workload-aware variants of depth-frequency cache compression. Although FreqDepthKV requires no retraining, models could be trained or lightly adapted to make adjacent-layer cache structure more compressible without weakening long-context reasoning \cite{gelberg2026training,cai2025redundancy}. The same framework could also be specialized for domains with distinct evidence patterns, such as tables, retrieval-augmented QA, multimodal inputs, and multi-agent inference \cite{corallo2025tablekv,silva2025raptor,zhang2025enhancing,patel2026polykv}. These directions would help determine when depth-frequency sharing should remain a purely inference-time system and when model- or workload-specific adaptation provides additional robustness.